\documentclass[runningheads]{llncs}
\usepackage{graphicx}
\usepackage{amsmath}
\usepackage{booktabs}
\usepackage{hyperref}
\usepackage{multirow}

\title{Perceptual Evaluation of GANs and Diffusion Models for Generating X-rays}
\titlerunning{Perceptual Evaluation of Generative X-ray Models}
\author{Gregory Schuit\inst{1,2,3}, Denis Parra\inst{1,2,3}\orcidID{0000-0001-9878-8761}, Cecilia Besa\inst{1,2}\orcidID{0000-0002-0015-0434}}

\authorrunning{G. Schuit et al.}

\institute{
Pontificia Universidad Católica de Chile, Chile\\
\and
iHealth -  Instituto Milenio en Ingeniería e Inteligencia Artificial para la Salud, Chile\\
\and
CENIA – Centro Nacional de Inteligencia Artificial, Chile\\
\email{\{gkschuit,dparras,cbesa\}@uc.cl}
}

\begin{document}

\maketitle

\begin{abstract}
Generative image models have achieved remarkable progress in both natural and medical imaging. In the medical context, these techniques offer a potential solution to data scarcity—especially for low-prevalence anomalies that impair the performance of AI‑driven diagnostic and segmentation tools. However, questions remain regarding the fidelity and clinical utility of synthetic images, since poor generation quality can undermine model generalizability and trust.

In this study, we evaluate the effectiveness of state-of-the-art generative models—Generative Adversarial Networks (GANs) and Diffusion Models (DMs)—for synthesizing chest X-rays conditioned on four abnormalities: Atelectasis (AT), Lung Opacity (LO), Pleural Effusion (PE), and Enlarged Cardiac Silhouette (ECS). Using a benchmark composed of real images from the MIMIC-CXR dataset and synthetic images from both GANs and DMs, we conducted a reader study with three radiologists of varied experience. Participants were asked to distinguish real from synthetic images and assess the consistency between visual features and the target abnormality.

Our results show that while DMs generate more visually realistic images overall, GANs can report better accuracy for specific conditions, such as absence of ECS. We further identify visual cues radiologists use to detect synthetic images, offering insights into the perceptual gaps in current models. These findings underscore the complementary strengths of GANs and DMs and point to the need for further refinement to ensure generative models can reliably augment training datasets for AI diagnostic systems.

\keywords{Medical Imaging \and Generative Models \and Chest X-rays \and Diffusion Models \and GANs}
\end{abstract}

\section{Introduction}

Recent advances in generative AI have led to remarkable progress in image synthesis across domains, including medical imaging. These models—particularly Generative Adversarial Networks (GANs) \cite{goodfellow2020generative} and Diffusion Models (DMs) \cite{rombach2022high}—have the potential to address longstanding challenges in medical AI, such as data scarcity for low-prevalence conditions and the generation of counterfactual explanations for model interpretability.

While Diffusion Models have recently surpassed GANs in image quality and are gaining popularity in chest X-ray (CXR) generation \cite{muller2212diffusion}, their clinical reliability remains underexplored. Most existing studies evaluate these models using proxy metrics like Frechet Inception Distance (FID) \cite{segal2021evaluating}, which do not capture radiological fidelity or diagnostic relevance. Although some works include expert feedback, they often rely on small image sets or overlook the reasons behind radiologists’ judgments.

To address this gap, we conduct a human-centered evaluation comparing GAN- and DM-generated chest X-rays. We focus on four common abnormalities: Atelectasis, Lung Opacity, Pleural Effusion, and Enlarged Cardiac Silhouette. Our study involves two structured user studies with three board-certified radiologists, who assess image realism and the presence of the intended pathology. We also collect qualitative insights on the visual cues used to identify synthetic content.

Our contributions are twofold: (1) we extend prior evaluation frameworks with a detailed, anomaly-specific analysis based on expert assessments, and (2) we offer a direct comparison between two state-of-the-art generative models, revealing trade-offs in realism versus conditional accuracy. Our findings highlight the complementary strengths of GANs and DMs and underscore the importance of human-in-the-loop validation in medical image synthesis.

To foster reproducibility, we release the source code\footnote{\url{https://github.com/gregschuit/radiologist-perceptual-eval-xrays}}, labeling interface, and generated dataset. This work aims to guide future development of trustworthy generative models for medical imaging applications.

\section{Related Work}

Generative models for medical imaging—especially GANs \cite{goodfellow2020generative} and Diffusion Models (DMs) \cite{rombach2022high}—have shown promising results. GANs train through adversarial feedback between a generator and a discriminator, while DMs iteratively denoise random noise into images, achieving high fidelity but often at greater computational cost \cite{dhariwal2021diffusion}.

In synthetic medical image evaluation, Segal et al. \cite{segal2021evaluating} introduced a dual approach using FID scores and expert radiologist assessments. Their findings revealed that although synthetic X-rays sometimes appeared realistic, they often lacked clinical reliability. Mertes et al. \cite{mertes2022ganterfactual} focused on explainability, and their user study revealed significant improvements in mental models, explanation satisfaction, trust, emotions, and self-efficacy compared to traditional saliency-based explanation methods like LIME and LRP. This work highlighted the importance of considering human factors in evaluating synthetic medical images.

Comparative evaluations have increasingly favored DMs. Muller-Franzes et al. \cite{muller2212diffusion} showed that DMs outperformed GANs on multiple datasets, including CheXpert, with markedly better FID, precision, and recall. Their Medfusion study established DMs as the new standard for diverse, high-quality generation across pathologies.

Domain adaptation remains a challenge. Chambon et al. \cite{chambon2022roentgen} addressed this by introducing RoentGen, a diffusion model fine-tuned on chest X-rays and radiology reports. They demonstrated improved image quality and representation of specific diseases, along with modest boosts in downstream task performance when using synthetic training data.

Despite progress, gaps persist. Metrics like FID lack clinical interpretability, multi-disease synthesis remains difficult, and evaluation protocols are not standardized. Moreover, few studies analyze how radiologists discern image authenticity or examine which pathologies are well represented.

Our study addresses these gaps through a dual radiologist-led evaluation of GAN and DM-generated X-rays, focusing on clinical realism, conditional accuracy, and visual patterns that inform human judgment.

\begin{figure}[tpb]
    \centering
    \includegraphics[width=0.6\linewidth]{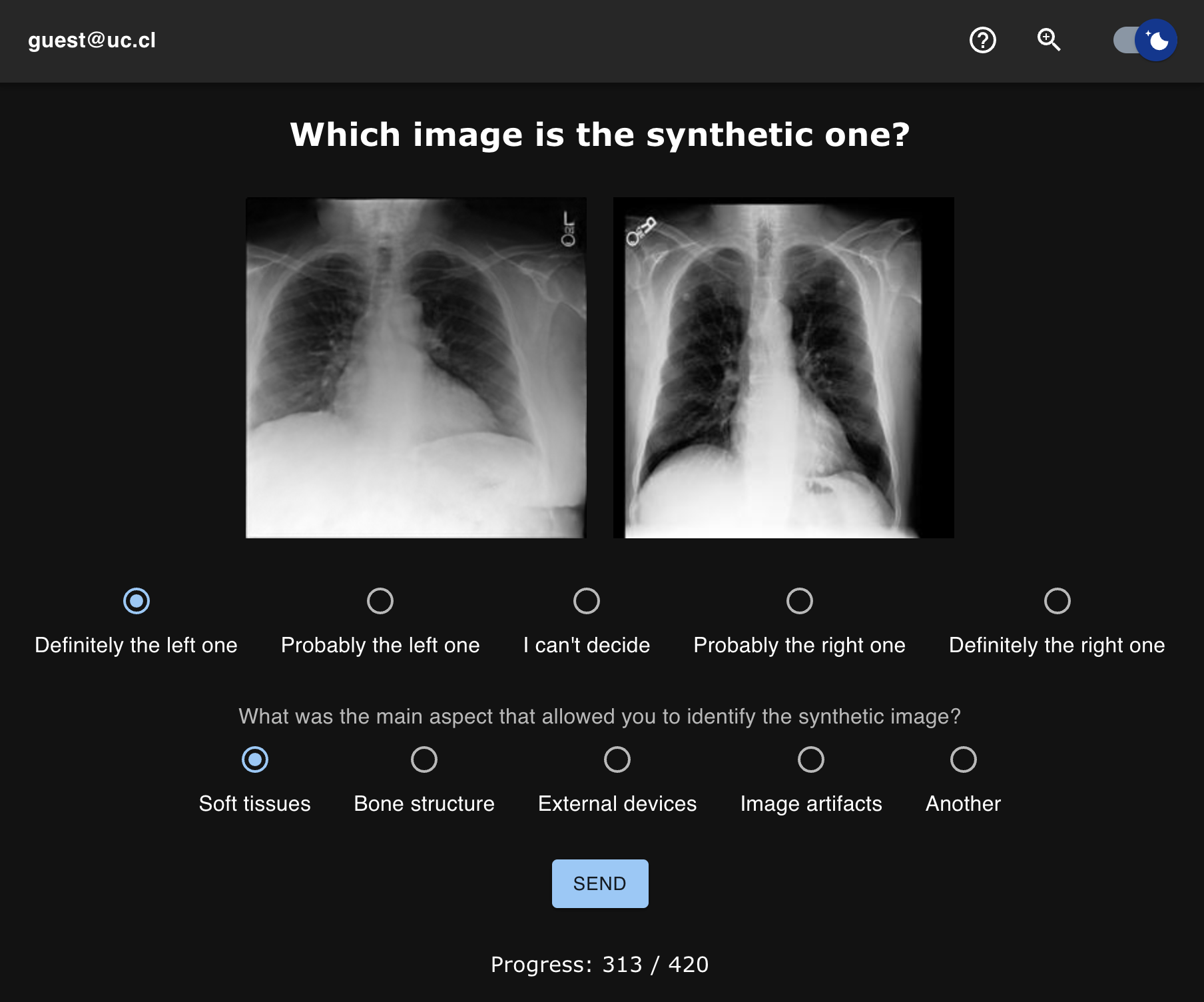}
    \caption{Interface of Task 1 concerning realism. The user is asked to identify which image is synthetic when displayed side-by-side with a real one. Five possible answers are displayed to capture different levels of confidence. Also, the user should select which aspect of the image influenced their decision.}
    \label{fig:web-interface-task-1}
\end{figure}

\section{Materials and Methods}

As mentioned in the introduction, it is essential to obtain radiologists' evaluations to ensure the validity of our generated images. Thus, we compiled a chest X-ray dataset that included real and synthetic images. Then, we engaged a group of three radiologists in two specific tasks. The first task focused on realism: radiologists were asked to identify the synthetic image when paired with a real one and to detail the particular features observed in each image. The second task involved assessing whether the image accurately represented the indicated condition. After that, we collected, processed, and analyzed the radiologists' labels to gauge the realism and conditional accuracy of the generated images. 

\begin{figure}[tpb]
    \centering
    \includegraphics[width=0.5\linewidth]{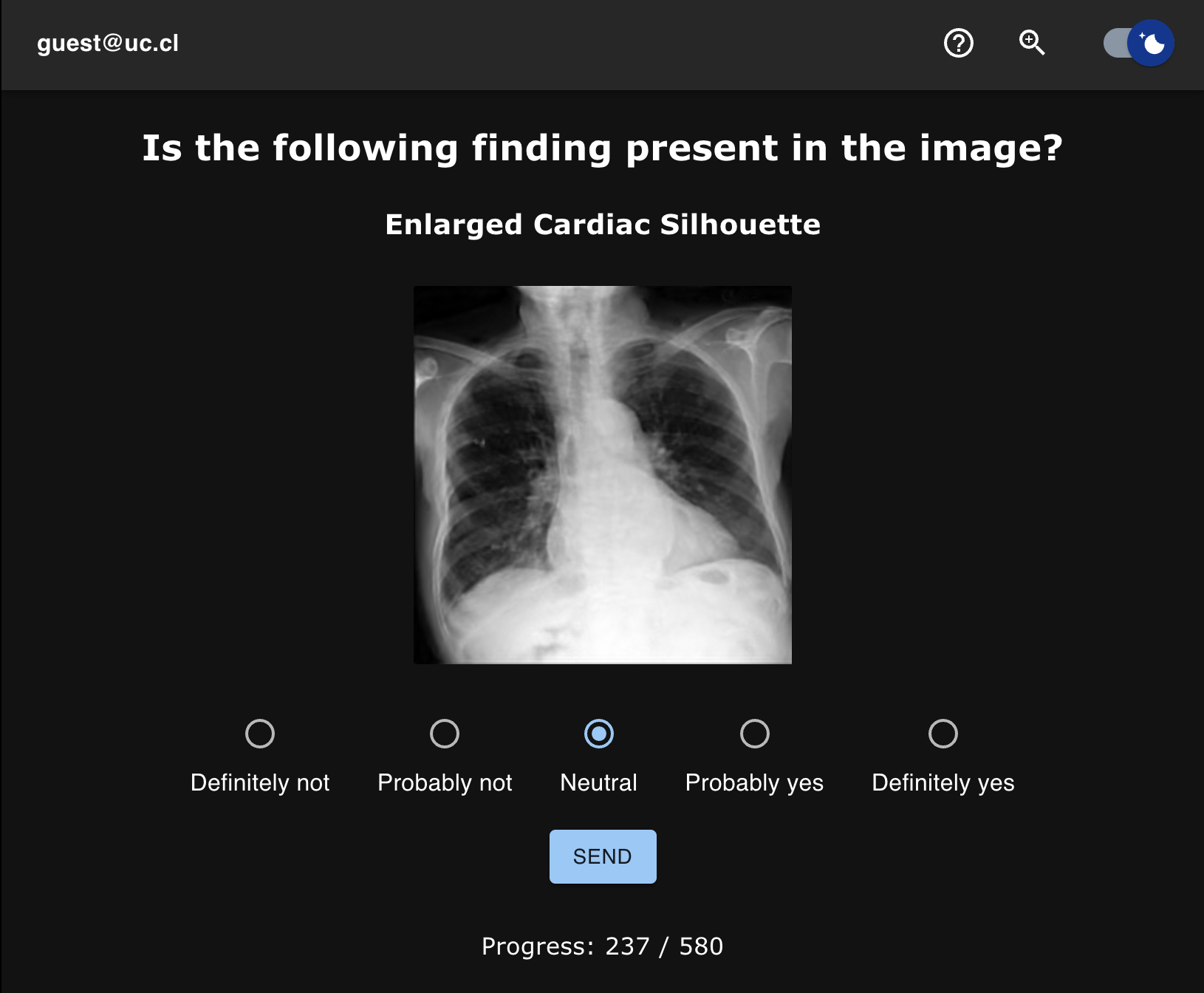}
    \caption{Web interface: Task 2 concerning conditionality. The user is asked whether the given label correctly classifies the displayed image. Both real and synthetic images are evaluated. The user should answer according to a scale of five levels of agreement.}
    \label{fig:web-interface-task-2}
\end{figure}

\subsection{Dataset Construction}
We created a dataset of 480 PA chest X-rays: 160 real images from the MIMIC-CXR \cite{johnson2019mimic} along with the Chest ImaGenome Dataset \cite{wu2021chest}, and 320 synthetic images (160 from a StyleGAN2 model \cite{hammer2022} and 160 from a text-conditioned Diffusion model for chest X-rays, RoentGen \cite{chambon2022roentgen}). The text to condition the X-ray generation on the presence of a finding was simply the name of the abnormality, e.g., \textit{Pleural Effusion}. To condition the generation for the absence of an abnormality, the input text was \textit{No} followed by the abnormality, e.g., \textit{No pleural effusion}.  Each image was associated with one of four abnormalities: Atelectasis (\textbf{AT}), Lung Opacity (\textbf{LO}), Pleural Effusion (\textbf{PE}), or Enlarged Cardiac Silhouette (\textbf{ECS}), marked as both present or absent. GAN images were generated with class-specific binary models; DM images were produced using RoentGen with text prompts.

\subsection{Evaluation Tasks}
Three radiologists identified as $Labeler_i$(L1: 13 yrs exp, L2: 1st-year resident, L3: 3rd-year resident) completed two tasks:

\textbf{Task 1: Realism Judgment} — Identify the synthetic image in a real-synthetic pair, with confidence level and feature-based justification (e.g., artifacts, devices). Figure \ref{fig:web-interface-task-1} shows the main interface on this task.

\textbf{Task 2: Conditionality Assessment} — Evaluate whether an image matched the abnormality label using a Likert scale, later binarized for accuracy metrics. Figure \ref{fig:web-interface-task-2} shows the main interface to collect radiologists' responses on this task.

\section{Results and Discussion} \label{sec:discussion}

\subsection{Are synthetic images discernible from real ones?}

In general, our results show that both models generate realistic images because most of the generated images are not discernible from real ones. We measured two metrics, Undecided Answers Rate (UAR), which indicates the proportion of cases in which the radiologists selected the neutral answer, i.e., they declared not to know which one was synthetic. Hence, higher values of UAR indicate higher model capability at realistic image generation. However, it is important to notice that lower values of UAR do not imply lower realism, as the outcome of the radiologist's decision could be wrong. We complement the analysis with the Correct Answer Rate (CAR) in those cases.

\begin{figure}[tpb]
    \centering
    \includegraphics[width=\linewidth]{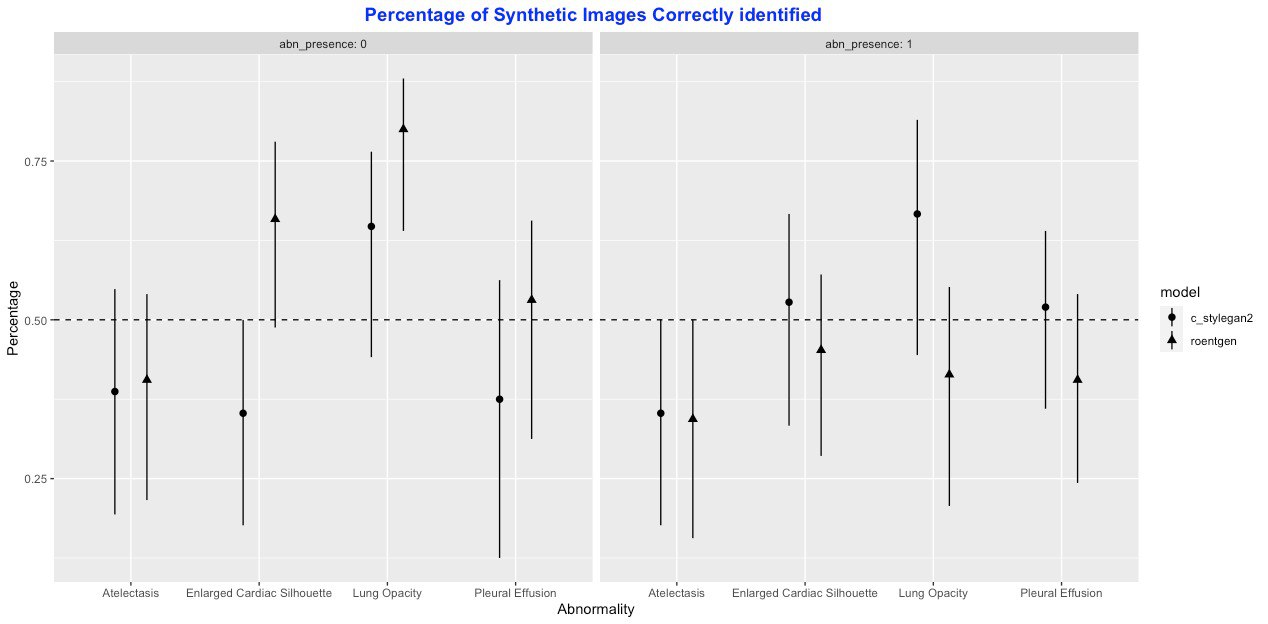}
    \caption{Percentage of Synthetic Images Correctly Identified by Human Labelers. Each bar in the plot indicates the corresponding bootstrapped confidence intervals.}
    \label{fig:car-confidence-intervals}
\end{figure}

Specifically, when identifying the synthetic image, radiologists only made a decision about which image was synthetic 58.5\% of the time, i.e., an UAR=41.8\%. Moreover, when a decision was made, they correctly recognized the synthetic image only 50.4\% of the time. This is consistent with the hypothesis that these models generate realistic enough images that are not discernible from real ones. However, some specific cases revealed poor realism, i.e., DM conditioned on the absence of LO (CAR = 80.0\%) and the absence of ECS (CAR = 65.9\%), as seen in Figure \ref{fig:car-confidence-intervals}. As we detail next, our posterior review with radiologists pointed out some concrete visual features that revealed the image's artificial nature in these conditions.

\subsection{What attributes can radiologists use to distinguish synthetic images?}

The posterior review of the reasons that radiologists used most as heuristics revealed that specific image attributes were successfully used for discrimination. Such attributes were: high radiolucency, incomplete pulmonary fields, abnormally large densities, and blurry lateral views.

\begin{figure}[htpb]
    \centering
    \includegraphics[width=0.7\linewidth]{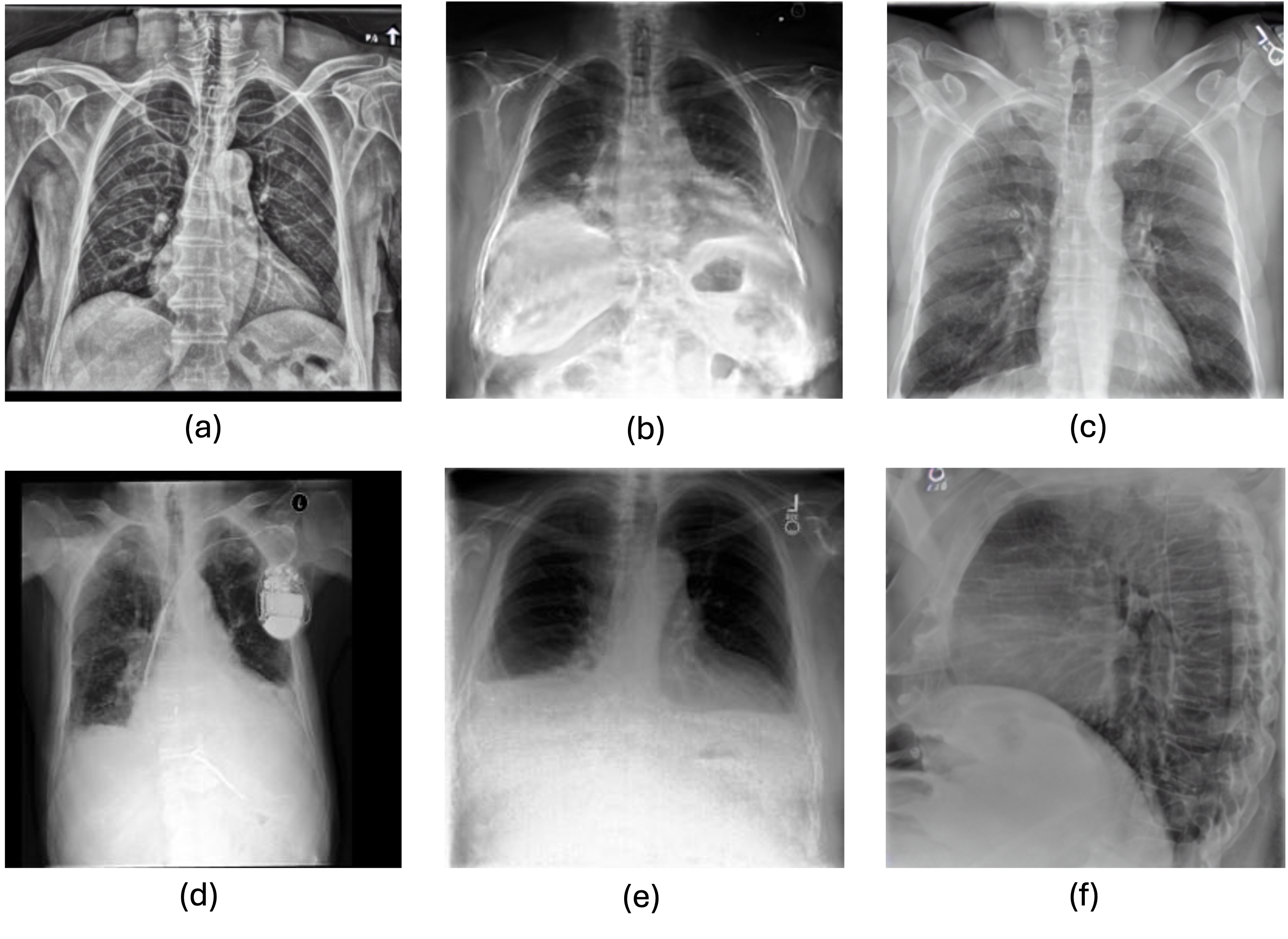}
    \caption{Representative examples of labelers' heuristics: a) extreme radiolucency, b) anomalous densities in it, c) image is cut, and the lung fields cannot be observed entirely, d) real image incorrectly selected as synthetic since it had a pacemaker, e) real image incorrectly selected as synthetic because of its poor technique, f) synthetic image that was correctly identified as such because it represents a blurry lateral view.}
    \label{fig:top-examples}
\end{figure}

Analyzing the visual patterns used in each case, we found that when they selected ``2. Bone structure", they used \textbf{high radiolucency as an effective heuristic} when dealing with DM-generated images without ECS. This same behavior arose when they selected the reason ``4. Image Artifacts." 

Similarly, they correctly used \textbf{cropped image as an effective heuristic} when dealing with DM-generated images without LO. Again, this heuristic was also used for the same case when the reason was ``4. Image artifacts." Further conversation with the participants revealed that an unclear definition of those reasons caused this. For example, as it can be seen in Figure \ref{fig:top-examples}, a high radiolucency can be interpreted as a bone structure property because bones are highlighted with the images' contrasts, and it can also be interpreted as a general artifact of the image. The same happens with the cropped images because if the image is incomplete, it can be associated with an incomplete bone structure or as a general image artifact. Also, this last visual feature was the most common aspect when they selected the free-text ``Other" option.

Another remarkable pattern in the radiologists' heuristics concerns \textbf{external devices}. Radiologists interpreted real images as synthetic when they had certain external devices, such as arthrodesis, catheters, pacemakers, and clips. \textbf{This constituted a wrong heuristic}. In the posterior review of these cases, we found that only the two less experienced radiologists followed this heuristic because they did not recognize the external devices. This suggests that participants' proper training must be ensured in future studies.

The reason \textbf{``Lung field - Cut image" was also a good heuristic}, especially when dealing with DM-generated images without LO. This pattern is important because it points to the main reason why radiologists got a higher CAR when dealing with DM-generated images without LO. This means that, when making a decision, radiologists were correct most of the time because they used this heuristic.

{\setlength{\tabcolsep}{0.5em} 

\begin{table*}[t]
\begin{center}
\tiny
\caption{Accuracy, True Positive Rate (Sensitivity), and True Negative Rate (Specificity) of radiologists evaluating consistency between images and labels.}\label{table:accuracies_tpr_and_tnr}
\vspace{1em}
\begin{tabular}{@{}llllllllllllll@{}}
\toprule
 &
   &
  \multicolumn{3}{l}{Accuracy} &
  \multicolumn{3}{l}{TPR} &
  \multicolumn{3}{l}{TNR} &
  \multirow{2}{*}{\begin{tabular}[c]{@{}l@{}}Mean\\ Acc.\end{tabular}} &
  \multirow{2}{*}{\begin{tabular}[c]{@{}l@{}}Mean\\ TPR\end{tabular}} &
  \multirow{2}{*}{\begin{tabular}[c]{@{}l@{}}Mean\\ TNR\end{tabular}} \\ \cmidrule(r){1-11}
                                      &      & L1    & L2    & L3    & L1     & L2     & L3     & L1     & L2     & L3     &      &       &      \\ \midrule
\multirow{3}{*}{AT}                 & real & .650 & .500 & .513 & .550  & .250  & .632  & .750  & .750  & .400  & .554 & .477  & .633 \\ \cmidrule(l){2-14} 
& GAN & .561 & .659 & .585 & .250  & .300  & .650  & .857  & 1.000 & .524  & .602 & .400  & .794 \\ \cmidrule(l){2-14} 
                                             & DM   & .600 & .550 & .550 & .950  & .900  & .900  & .250  & .200  & .200  & .567 & .917  & .217 \\ \midrule
\multirow{3}{*}{ECS} & real & .763 & .789 & .789 & .850  & .800  & .900  & .667  & .778  & .667  & .781 & .850  & .704 \\ \cmidrule(l){2-14} 
& GAN & .875 & .875 & .850 & .950  & .950  & .900  & .800  & .800  & .800  & .867 & .933  & .800 \\ \cmidrule(l){2-14} 
                                             & DM   & .875 & .700 & .718 & 1.000 & 1.000 & 1.000 & .750  & .400  & .450  & .764 & 1.000 & .533 \\ \midrule
\multirow{3}{*}{LO}                & real & .675 & .600 & .750 & .400  & .200  & .650  & .950  & 1.000 & .850  & .675 & .417  & .933 \\ \cmidrule(l){2-14} 
& GAN & .553 & .553 & .474 & .056  & .111  & .056  & 1.000 & .950  & .850  & .526 & .074  & .933 \\ \cmidrule(l){2-14} 
                                             & DM   & .450 & .436 & .450 & .150  & .263  & .400  & .750  & .600  & .500  & .445 & .271  & .617 \\ \midrule
\multirow{3}{*}{PE}           & real & .800 & .825 & .850 & .650  & .650  & .800  & .950  & 1.000 & .900  & .825 & .700  & .950 \\ \cmidrule(l){2-14} 
& GAN & .868 & .615 & .842 & .810  & .318  & .714  & .941  & 1.000 & 1.000 & .775 & .614  & .980 \\ \cmidrule(l){2-14} 
                                             & DM   & .900 & .900 & .923 & .800  & .800  & .900  & 1.000 & 1.000 & .947  & .908 & .833  & .982 \\ \bottomrule
\end{tabular}
\end{center}
\end{table*}

}

\subsection{Are generative models effective at generating conditioned abnormalities?}

The results regarding the models' ability to follow the requested conditionality were mixed. Although ECS and PE were generated with high accuracy in both models, LO and AT presented challenges, as seen in Table \ref{table:accuracies_tpr_and_tnr}.

Specifically, results regarding the accurate generation of LO point to two gaps. The first one is the bad performance of radiologists in front of real images. Presumably, this is due to the limitations of the image format. The resolution used in the interface and during the training of GAN was 256x256 pixels, and it is very low compared to the DICOM format used by radiologists in a real diagnosis setup, which is usually above a thousand pixels square. Also, radiologists pointed out that contrast is a critical aspect that can be regulated when they inspect real radiographs, and it is also a key attribute when spotting lung opacities. This could explain why the three labelers performed badly at classifying LO within the real image set, especially in terms of TPR, i.e., in the cases where the abnormality was supposed to be present. Moreover, we see high values of TNR. This reflects a tendency to answer that there was no LO.

However, labelers performed even worse when confronting model-generated images conditioned on LO. This brings us to the second gap: the model's lack of efficacy in generating LO. As the results show, TPR is notably lower than the TNR in a more pronounced way in the synthetic sets than in the real set. This implies that, aside from the limitation posed by radiologists tagging format-limited images, there is also a performance issue with the generative models. Although this aspect was not studied in detail, we suspect critical information for generating lung opacity contrasts may be lost during the model training process. Further investigation in this line is left as future work.

\subsection{Are DMs superior to GANs at generating chest X-rays?}

Regarding realism, no significant differences were found in most categories when comparing GAN and DM. However, the GAN-based model showed significantly superior performance when conditioned on the absence of ECS. As explored in the radiologists' heuristics analysis, this difference is mainly due to the high radiolucency present in some DM-generated images when the model is prompted with no ECS. However, the reasons behind this behavior are unclear and have not been further explored in this study.

Overall, GAN demonstrated slightly better performance regarding conditional correctness compared to the DM. This suggests that, despite the DM's advantage of accepting natural language descriptions as conditions, the GAN-based model might be more effective at generating synthetic images with precise binary conditionality. However, this is subject to the prompts used with the DM. Future work should study whether changing the prompting of a text-condition generation can improve the results for DM models observed in this study. 

\subsection{Limitations}

Our findings should be interpreted with caution due to several limitations. First, the relatively small sample size impacts the statistical power and generalizability of the results. Additionally, the limited number of participants may restrict the breadth of insights derived from the data. Expanding the sample size and the participant pool in future studies could reveal significant undetected differences in the current dataset. Besides, utilizing only one prompt template for the Diffusion Model represents a limitation. The model prompts can dramatically change its behavior, and different prompting techniques could yield different image quality. Another limitation is the small size of the images used in the study (256x256), which decreases how representative the experiment is of a clinical scenario using DICOM images.

\section{Conclusion}

Since generative AI has shown impressive results in the general domain, it is tempting to believe that it could solve crucial problems in medical imaging, like data scarcity, privacy, and explainability. In addition, DMs have gained popularity due to their ability to generate high-quality images conditioned on free text, and some studies show that DMs beat GANs in the general domain \cite{muller2212diffusion}. However, our results constitute evidence that contradicts these assertions in the medical domain. Specifically, our results show that the generation of realistic chest radiographs is not yet a solved problem, because there are clear situations in which these images are suboptimal and recognizable.

\section*{Acknowledgments}

This work was supported by ANID Chile, the Millennium Science Initiative Program, code ICN2021\_004 (iHealth), by Chile's National Center for Artificial Intelligence, Basal Fund code FB210017 (CENIA), and by Fondecyt Regular grant 1231724.
 
\bibliographystyle{splncs04}
\bibliography{Paper-0016}
\end{document}